\documentclass[conference]{IEEEtran}
\IEEEoverridecommandlockouts
\usepackage{cite}
\usepackage{amsmath,amssymb,amsfonts}
\usepackage{algorithmic}
\usepackage{graphicx}
\usepackage{textcomp}
\usepackage{xcolor}
\usepackage{setspace}
\usepackage[colorlinks,
            linkcolor=red,
            anchorcolor=blue,
            citecolor=green
            ]{hyperref}
\def\BibTeX{{\rm B\kern-.05em{\sc i\kern-.025em b}\kern-.08em
    T\kern-.1667em\lower.7ex\hbox{E}\kern-.125emX}}

\begin{document}

\title{\textbf{\LARGE{Tell Codec What Worth Compressing: Semantically Disentangled Image Coding for Machine with LMMs}}\vspace{-6mm}}
\author{
\normalsize{Jinming Liu$^1$$^,$$^2$, Yuntao Wei$^2$, Junyan Lin$^2$, Shengyang Zhao$^2$, Heming Sun$^4$, Zhibo Chen$^3$, Wenjun Zeng$^2$, Xin Jin$^{2\dag}$}\\
\normalsize{$^1$Shanghai Jiao Tong University \qquad
$^2$Ningbo Institute of Digital Twin, Eastern Institute of Technology, Ningbo, China}\\
\normalsize{$^3$University of Science and Technology of China\qquad
$^4$Yokohama National University\vspace{-5mm}}}

\maketitle
\begin{abstract}
We present a new image compression paradigm to achieve ``intelligently coding for machine'' by cleverly leveraging the common sense of Large Multimodal Models (LMMs). We are motivated by the evidence that large language/multimodal models are powerful general-purpose semantics predictors for understanding the real world. Different from traditional image compression typically optimized for human eyes, the image coding for machines (ICM) framework we focus on requires the compressed bitstream to more comply with different downstream intelligent analysis tasks. To this end, we employ LMM to \textcolor{red}{tell codec what to compress}: 1) first utilize the powerful semantic understanding capability of LMMs w.r.t object grounding, identification, and importance ranking via prompts, to disentangle image content before compression, 2) and then based on these semantic priors we accordingly encode and transmit objects of the image in order with a structured bitstream. In this way, diverse vision benchmarks including image classification, object detection, instance segmentation, etc., can be well supported with such a semantically structured bitstream. We dub our method ``\textit{SDComp}'' for ``\textit{S}emantically \textit{D}isentangled \textit{Comp}ression'', and compare it with state-of-the-art codecs on a wide variety of different vision tasks. SDComp codec leads to more flexible reconstruction results, promised decoded visual quality, and a more generic/satisfactory intelligent task-supporting ability.  
\end{abstract}


\vspace{-2.5mm}
\section{Introduction}
\vspace{-1mm}
As the foundation for visual signals communication~\cite{wallace1992jpeg}, image compression effectively reduces the storage space and transmission bandwidth required without largely compromising their quality. With the recent advancements in large model-based machine vision tasks like image classification~\cite{radford2021learning}, object detection~\cite{ren2015faster}, and segmentation~\cite{kirillov2023segment}, the integration of substantial and extensive machine task data into communication protocols has posed more challenges due to their intolerable storage resource requirements. This scenario has made machine-oriented image compression a challenging and demanding research topic, which compresses images to a bitstream that is more efficient and task-friendly. 

However, despite having demonstrated strong rate-distortion performance, traditional image compression methods, including JPEG~\cite{wallace1992jpeg}, HEVC~\cite{sullivan2012overview}, and VVC~\cite{vvc}, primarily focus on visual fidelity and realism for meeting human perception~\cite{Liu_2023_CVPR,agustsson2023multi,careil2023towards}, while neglecting machine perception, thus result in a suboptimal supporting to the downstream machine tasks~\cite{duan2020video}.

\begin{figure}
    \centering
    \includegraphics[width=\linewidth]{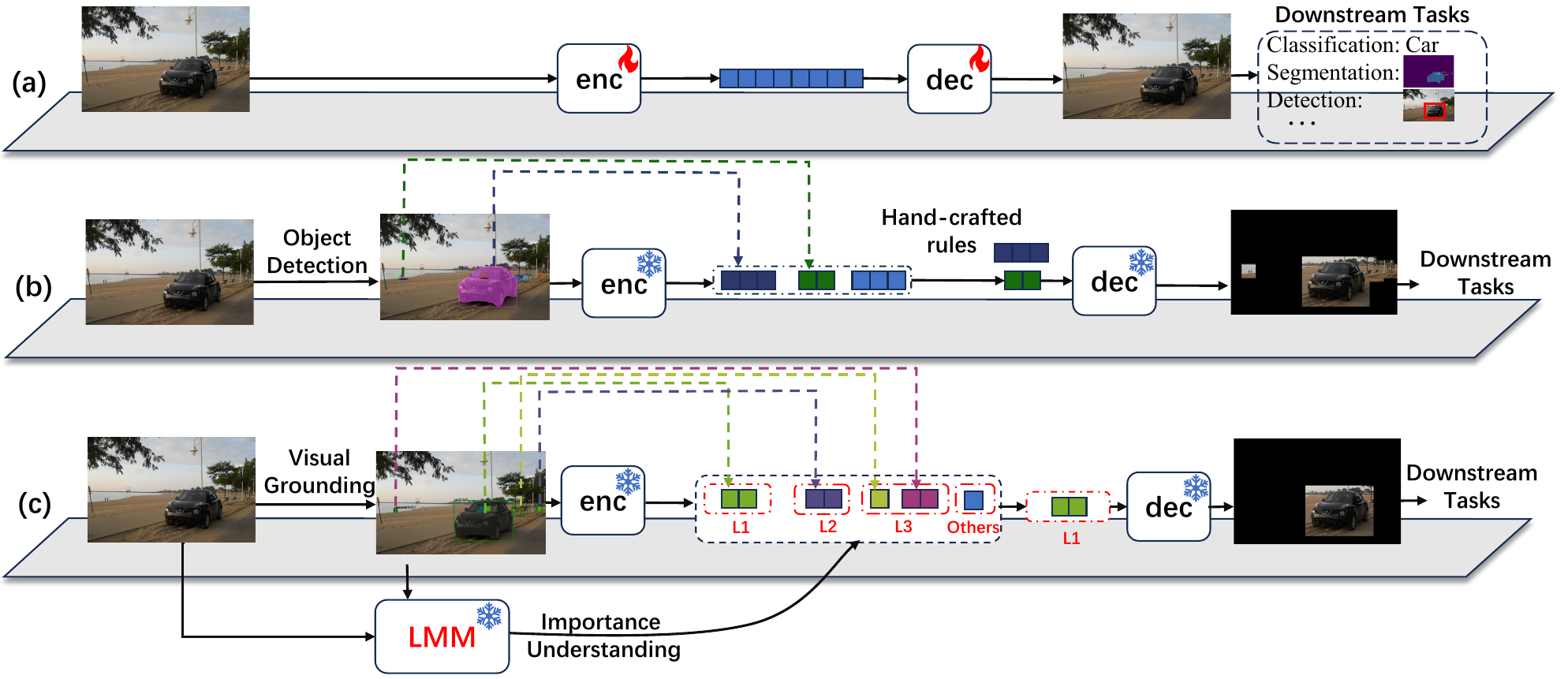}
    \vspace{-8.5mm}
    \caption{Illustration of (a) Tasks-driven ICM framework, (b) Semantically structured image compression and hand-crafted regions selection, (c) Our SDComp framework driven by LMM. SDComp employs visual-grounding to structure the image by dividing it into distinct regions. These regions are then evaluated for importance using a Large Multimodal Model (LMM). Based on their importance, the regions are encoded and transmitted sequentially.}
    \vspace{-4mm}
    \label{fig:teaser}
    \vspace{-2mm}
\end{figure}

To address this issue, a new compression branch of image coding for machines (ICM) has been proposed in recent years. This paradigm aims to optimize codecs for downstream machine vision tasks.
For instance, Chamain \textit{et al.} ~\cite{chamain2021end} proposed a joint optimization of codecs and downstream task networks to enhance machine vision tasks. 
Feng \textit{et al.} ~\cite{Feng_2023_ICCV} introduced a structured bitstream approach, which improves downstream tasks by transmitting only the specified objects. However, as shown in Figure~\ref{fig:teaser}, these methods at least have the following limitations: 1). Less flexible — 
the codec requires retraining the compression network for different tasks to select task-driven features/regions for transmission; 
2). Poor practicality — users have to know the type of downstream tasks on the decoding side to design hand-crafted rules for selecting task-related features/regions for bitstream transmission;
3). Low efficiency for machine — weak semantics understanding capability makes codec lack of foucs.

To overcome these obstacles, \textbf{the key is how to help codec figure out what is really important for downstream tasks, i.e., what is more worth to be compressed?} So, in this paper, a large multimodal model (LMM) is used to assist in understanding the image and rank the importance of objects within the image, which in advance captions the image semantics before compression, leading to a semantics-clear and disentangled coding process. In this way, a semantically structured bitstream can be generated, and then transmitted sequentially according to the importance ranking, which allows direct operation at the bitstream level to support different tasks such as classification, detection, etc, achieving an interpretable and controllable downstream task supporting. The feasibility of such a design is promised by the fact that large language models (LLM) with scaling law~\cite{kaplan2020scaling} have experienced rapid development in recent years. They have demonstrated various emergent abilities~\cite{brown2020language,wei2022emergent}.
Moreover, with the advancement of LLMs, LMMs have also made significant progress, e.g., LLaVA~\cite{liu2024visual} and MiniGPT-4~\cite{zhu2023minigpt} demonstrate impressive results in instruction-following and visual reasoning capabilities.

In short, we are exploring a novel compression paradigm dubbed \textit{Semantically Disentangled Compression (SDComp)}. It firstly utilizes grounded-SAM~\cite{ren2024grounded,liu2023grounding,kirillov2023segany} to locate objects within the image. Compared with object detection methods~\cite{Feng_2023_ICCV,ren2015faster} that extract instances from images, visual grounding offers stronger zero-shot capabilities, enabling the detection of instances with different granularities and those that have not appeared in the training data. Subsequently, we fully leverage the understanding capabilities of LMM~\cite{chen2024far} by purposely designing prompts to rank these instances by importance. Finally, we leverage a method of Semantically Structured Image Compression (SSIC) method~\cite{Feng_2023_ICCV,sun2020semantic} and upgrade it with the semantic priors brought by LMM, to sequentially encode and transmit these instances in order of their importance. When facing downstream tasks, we can achieve excellent task performance by transmitting only the important parts of bitstreams.
Our contributions can be summarized as follows:
\begin{itemize}

    \item It is the first to utilize large multimodal models (LMMs) to improve the performance of image coding for machines. The semantics information like object localization and image captions are used as query prompts for LMMs to perform importance ranking and semantic coding.
    
    \item Technically, our method proposes a prompt design rule for using LMM well, and designs a prior-embedding semantically disentangled compression framework, which allows our codec to be adapted to various downstream tasks flexibly with task-relevant bitstreams, demonstrating strong generalization capabilities.
    \item Experiments have proven the effectiveness of our approach, achieving an average performance improvement of 32.3\% over VTM on the segmentation, detection tasks. Additionally, experiments on visual question-answering tasks further validate the generalization of our method. We also explore the interpretability of our approach through both prompt-based and visualization techniques.
\end{itemize}

\vspace{-2mm}
\section{Related Works}
\vspace{-1mm}
\subsection{Image Coding for Machine}
\vspace{-1mm}
Traditional classical image/video compression standards, e.g., JPEG~\cite{wallace1992jpeg}, AV1~\cite{han2021technical}, HEVC~\cite{sullivan2012overview}, VVC~\cite{vvc}, and learning-based schemes~\cite{cheng2020learned,Liu_2023_CVPR,balle2017end,balle2018variational,lu2019dvc}, have been fully explored and widely deployed to satisfy human perception, and achieve significant performance on fidelity metrics such as PSNR and MS-SSIM.
However, the compression artifacts caused by these methods still significantly downgrade the performance of downstream intelligent visual tasks like object detection.

Therefore, 
a new compression diagram emerges, specifically for machine analysis, dubbed Image Coding for Machines (ICM)~\cite{he2019beyond,duan2020video,sun2020semantic,jin2023semantical,liu2022semantic,liu2023composable}, which aims to enhance the performance of downstream vision tasks such as detection~\cite{ren2015faster}, and segmentation~\cite{chen2017deeplab} by optimizing image codec. 
Current ICM approaches can be categorized into two tracks: 1). compression followed by analysis—joint optimization of image codecs and task-specific models or 2). analysis followed by compression—select partial regions/features that are task-interested for compression instead of the entire image~\cite{liu2021learning}. For example, Chamain $et\ al.$ ~\cite{chamain2021end} investigated the joint optimization of codecs and task networks to improve performance on specific tasks. Feng $et\ al.$ ~\cite{Feng_2023_ICCV} proposed a structured compression approach that encodes only the detected objects for machine analysis. However, current ICM methods are typically limited by task-driven optimizations or require hand-crafted designs, which hinders codec from generalizing to wider machine tasks. Besides, the existing ICM codecs lack good semantic understanding capability and hardly support machine intelligence purposely in a flexible human-like manner.

\vspace{-2mm}
\subsection{Visual Grounding for Semantics Perception}
\vspace{-1mm}
Visual grounding~\cite{karpathy2015deep,ren2024grounded,liu2023grounding} is a critical research area in computer vision and natural language processing, involving the alignment of textual descriptions with specific regions in an image. These methods~\cite{liu2019learning,wang2019neighbourhood,yang2019dynamic} typically show powerful reasoning and understanding capabilities to perform tasks such as object localization and referring expression comprehension~\cite{sadhu2019zero,yang2020improving,yang2019fast}. Such powerful fine-grained cross-modal perception capabilities are exactly what machine-oriented codecs need, so we consider empowering our method with visual grounding priors to achieve ``intelligently coding for machine''.

\vspace{-2mm}
\subsection{Large Multimodal Model for Semantics Understanding}
\vspace{-1mm}
Large Multimodal Models (LMM) bring their powerful reasoning and understanding capabilities into multimodal tasks, enabling many visual tasks, such as visual question answering \cite{yang2023dawn,liu2024visual,you2023ferret}, and document reasoning \cite{hong2024cogagent,liu2024textmonkey}. Representatively, BLIP-2 \cite{li2023blip} introduced Q-Former acting as a bottleneck between the visual encoder and the LLM. LLaVA \cite{liu2024visual} constructed a visual instruction-following dataset and achieved robust performance using a simple linear projector to align the image and text domains. 
These studies indicate that LMMs exhibit strong semantics understanding abilities, so we intend to utilize these semantic priors to accordingly compress and transmit objects of the image purposely, to support more downstream tasks in a flexible human-like manner.

\begin{figure*}[ht]
    \centering
    \includegraphics[width=0.98\linewidth]{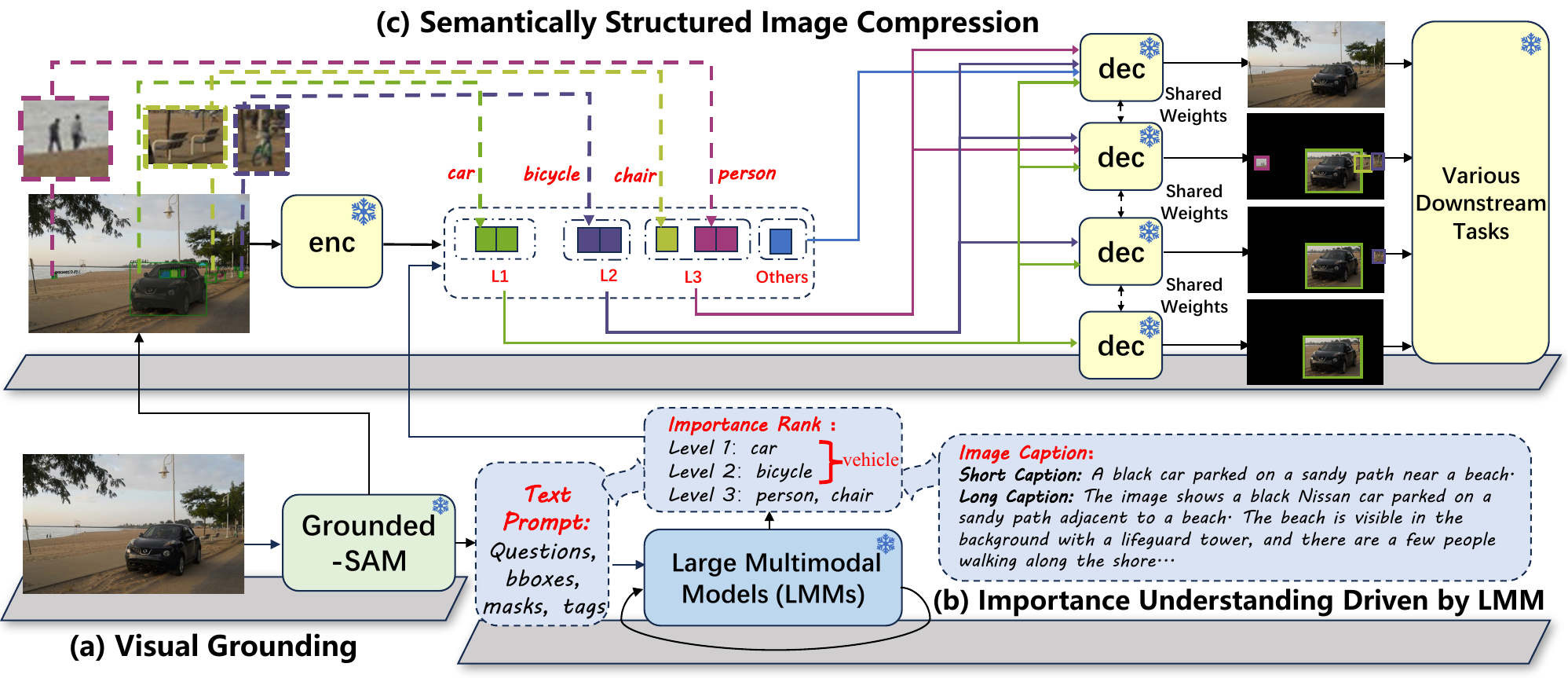}
    \vspace{-4mm}
    \caption{Overall framework of SDComp. (a) Grounded-SAM first extracts object grounding information for an image. (b) Such priors along with the designed prompts and generated captions instruct LMM to rank objects' importance, serving as a basis for (c) Semantically Structured Image Compression.}
    \vspace{-5mm}
    \label{fig:framework}
\end{figure*}
\vspace{-2mm}
\section{Semantically Disentangled Compression}
\vspace{-1mm}
\subsection{Framework Overview}
\vspace{-1mm}
\textbf{As we all know, the understanding of an image in human vision is usually progressive. Our eyes tend to first capture the most prominent objects in the image, then other secondary objects, and finally the background.} Inspired by this biological phenomenon, we design a compression paradigm dubbed Semantically Disentangled Compression (SDComp) to imitate this progressive importance comprehension, as shown in Figure \ref{fig:framework}. Given an image, we first use \textit{Visual Grounding} techniques to locate objects of the image, obtaining the positional information of different objects (bounding boxes and masks). Then, we instruct LMM to generate \textit{a pair of short\&long captions} for a coarse understanding of the image. Furthermore, we use these captions along with the previously extracted positional information and the image as inputs to design prompts that guide the LMM to \textit{rank the importance of different objects}. Finally, we upgrade \textit{semantically structured image compression (SSIC)}, encoding and transmitting objects in order of their importance. In this way, we can purposely control the transmission of objects based on their importance level as needed, saving bitrates for task-irrelevant information.

\subsection{Visual Grounding for Object Location}
\begin{figure}
    \centering
    \includegraphics[width=\linewidth]{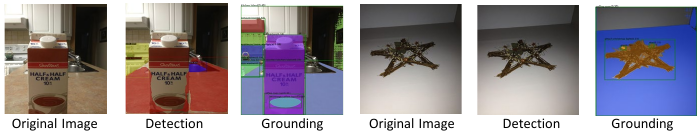}
    \vspace{-7.5mm}
    \caption{Comparison of detection and visual grounding. Compared with object detection, visual grounding can recognize open-vocabulary categories.}
    \vspace{-5mm}
    \label{fig:grounding_res}
\end{figure}

\vspace{-1mm}
Object detection networks typically detect objects of fixed categories~\cite{ren2015faster}, making it challenging to handle open-vocabulary scenarios. In contrast, visual grounding~\cite{ren2024grounded,liu2023grounding,kirillov2023segany}, which localizes objects based on text queries, offers better zero-shot and generalization capabilities, as shown in Figure~\ref{fig:grounding_res}.

Therefore, to enable our SDComp codec to efficiently and accurately understand the image in the wild world and provide better prior for ``intelligently coding for machine'', we adopt Grounded-SAM~\cite{ren2024grounded}, a kind of Open-Vocabulary detection and segmentation method for object location. Given an input image $x$ and a text prompt from manual queries or tags generated by RAM~\cite{zhang2023recognize} $M_{\text{RAM}}$, we first employ Grounding DINO~\cite{liu2023grounding} $M_{\text{G}}$ to generate precise boxes for objects or regions within the image to align the input textual information.
Subsequently, the annotated boxes obtained by Grounding DINO serve as the box prompts for SAM~\cite{kirillov2023segment} to generate precise mask annotations.
Then, these outputs are sent to LMM for objects' importance ranking within an image.
\vspace{-2mm}
\subsection{Importance Understanding by Large Multimodal Model}
\vspace{-1mm}
\begin{figure}
    \centering
    \includegraphics[width=\linewidth]{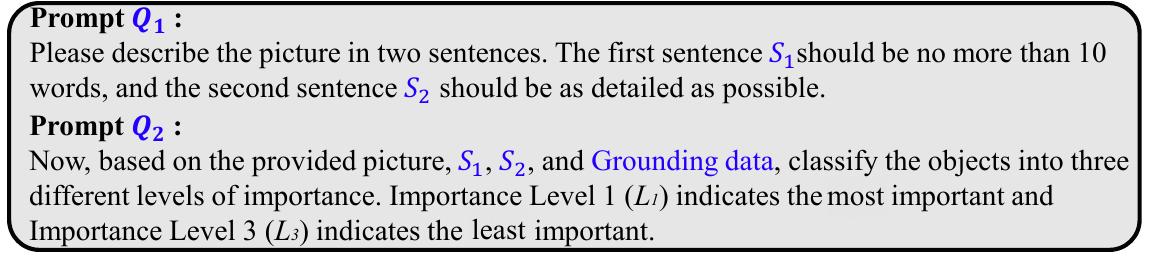}
    \vspace{-7.5mm}
    \caption{\textcolor{blue}{Q1:}The prompt template for generating short and long captions. \textcolor{blue}{Q2:}The prompt template for ranking the importance of objects, where \textcolor[HTML]{1d02f5}{Grounding data} contains the label, ID, and bounding box information.}
    \vspace{-5mm}
    \label{fig:template1}
\end{figure}

Accurately ranking the importance of objects within the image is crucial for the performance of intelligent tasks. Therefore, we utilized an advanced model, InternVL-Chat-V1.5 (InternVL 1.5 for short)~\cite{chen2023internvl}, in our SDComp for image understanding and reasoning, 
InternVL 1.5 possesses strong cross-modal semantic understanding and reasoning capabilities, which could help our codec work well in various machine tasks such as image captioning~\cite{li2023blip}, optical character recognition (OCR)~\cite{rang2023empirical} for detailed image perception, and visual question answering (VQA)~\cite{gurari2018vizwiz} for complex problem understanding. Furthermore, to fully unleash InternVL 1.5's ability to rank the importance of objects in images, we design two progressive prompts to guide the model step by step. Specifically, as shown in Figure \ref{fig:template1}, we first pose question $Q_1$ to InternVL 1.5, asking the model to provide a simple caption $S_1$ and a detailed caption $S_2$ for the image, respectively. Then, based on $S_1$, $S_2$ and the output of Grounded-SAM (including object labels, IDs, and bounding boxes), we rank the importance of the objects into three levels for a complexity-performance balance with prompts $Q_2$ shown in Figure \ref{fig:template1}.

\vspace{-2.mm}
\subsection{Semantically Structured Image Compression}
\vspace{-1mm}

The original SSIC~\cite{sun2020semantic} aims to compress specific local regions rather than the entire image information, which simply divides the image into objects and backgrounds. Thanks to the power of large models, we have developed a novel advanced SSIC architecture in our SDComp based on semantic priors. Specifically, we first structure the image as a combination of multiple objects and backgrounds $\{R_{obj1}, R_{obj2},..., R_{objn}, R_{back}\}$ based on the bounding box/mask obtained from visual grounding, following the setting in~\cite{Feng_2023_ICCV}. Subsequently, using the importance ranking derived from the LMM (Large Multimodal Model) as a prior, we group and sort the different structured regions according to their importance, dividing them into three important groups and an other group. Each important group contains the corresponding objects $\{R_{k_{obj1}}, R_{k_{obj2}},..., R_{k_{objm}}\}$ from each importance level, where the importance level $L_k$ ($k\in[1,3]$) contains $m$ objects. The remaining objects and the background are assigned to the other group, as less relevant parts for tasks.

\vspace{-1mm}
\section{Experiments}
\vspace{-1mm}
\subsection{Experiment Setting}
\vspace{-1mm}
\noindent\textbf{Model Setting.} For visual grounding, we utilize Grounded-SAM~\cite{ren2024grounded} to obtain localization information within the image. We employ InternVL 1.5~\cite{chen2023internvl}, which is one of the best open-source LMM available, to aid in image understanding. Finally, we use SSIC based on ELIC~\cite{he2022elic} to compress the image.

\noindent\textbf{Evaluation Protocol.}
Rate-distortion is used to evaluate SDComp, where distortion is defined as the metric corresponding to each machine vision task, and the rate is measured by bits per pixel (bpp).
We conduct experiments on various tasks including object detection, instance segmentation, image classification, visual question answering, and image captioning.

\noindent\textbf{Comparison Approaches.} 
The uncompressed images are input into the downstream task model to generate an upper-bound baseline. SDComp is compared with the mainstream image compression standard VVC (VTM)~\cite{vvc} and the learned codec ELIC~\cite{he2022elic}. Additionally, we also compare SDComp with ICM methods, SSIC~\cite{sun2020semantic} and ICMH-Net~\cite{liu2023icmh}.

\vspace{-2mm}
\subsection{Performance Comparison of Coding for Machine}
\vspace{-1mm}
\begin{figure}
    \centering
    \includegraphics[width=\linewidth]{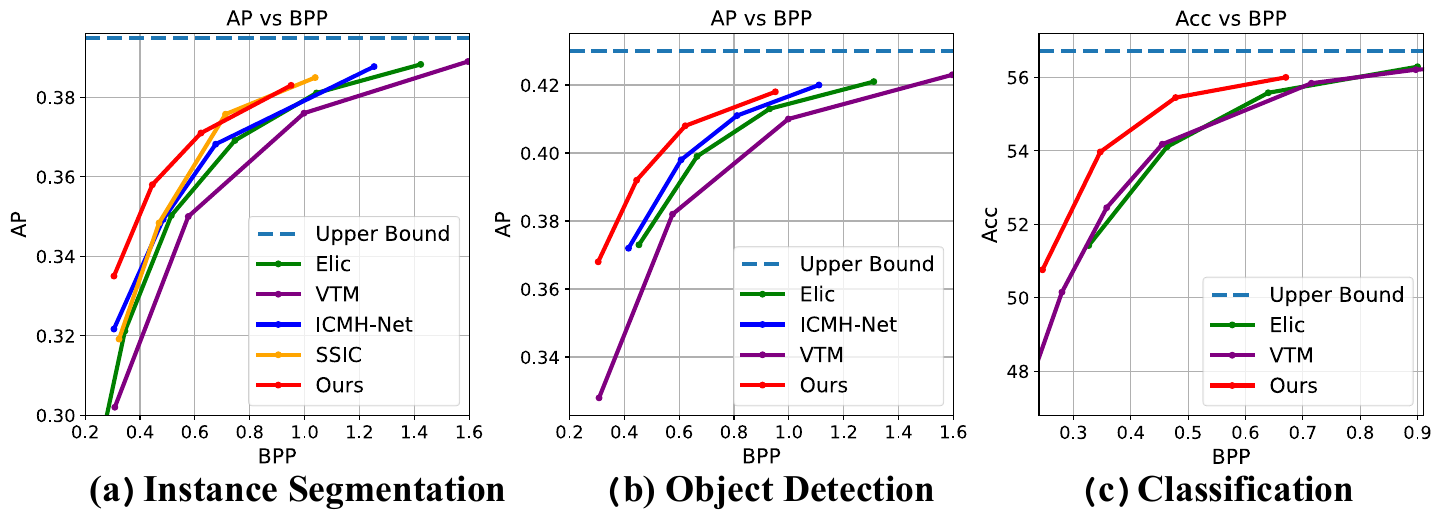}
    \vspace{-8mm}
    \caption{Performance Comparison of (a) Segmentation on COCO dataset. (b) Detection on COCO dataset. (c) Classification on CUB-200-2011 dataset.}
    \vspace{-4mm}
    \label{fig:performance}
\end{figure}
We first evaluate our method on three tasks: instance segmentation, object detection, and image classification. For segmentation and detection, we use Mask R-CNN~\cite{he2017mask} and Faster R-CNN~\cite{ren2015faster} (X101-FPN backbone) respectively, tested on the COCO dataset. For the classification task, we evaluate our method on ResNet-50 based on the CUB-200-2011 dataset~\cite{wah2011caltech}, which is a fine-grained classification dataset.

As shown in Figure~\ref{fig:performance}, our method demonstrates significant advantages in rate-distortion performance compared with the learned method ELIC~\cite{he2022elic} and the image compression standard VVC (VTM)~\cite{vvc} because of partially compression. In segmentation, detection, and classification tasks, SDComp can achieve BD-rate~\cite{Bjntegaard2001CalculationOA} reductions of -31.38\%, -33.22\%, and -12.83\% compared to VVC, respectively. 
Note that, SDComp just depends on $L_1$ to achieve the best performance for classification, which proves that partially decoding with main semantic content is sufficient to support some machine tasks. For dense visual tasks like segmentation, SDComp also achieves best by discarding task-irrelevant backgrounds.
Furthermore, we compare our method with the recent ICM methods~\cite{sun2020semantic,liu2023icmh}, and our approach also achieved superior performance on these tasks. 

\vspace{-2mm}
\subsection{Validation of Importance Understanding}
\begin{figure}
    \centering
    \includegraphics[width=\linewidth]{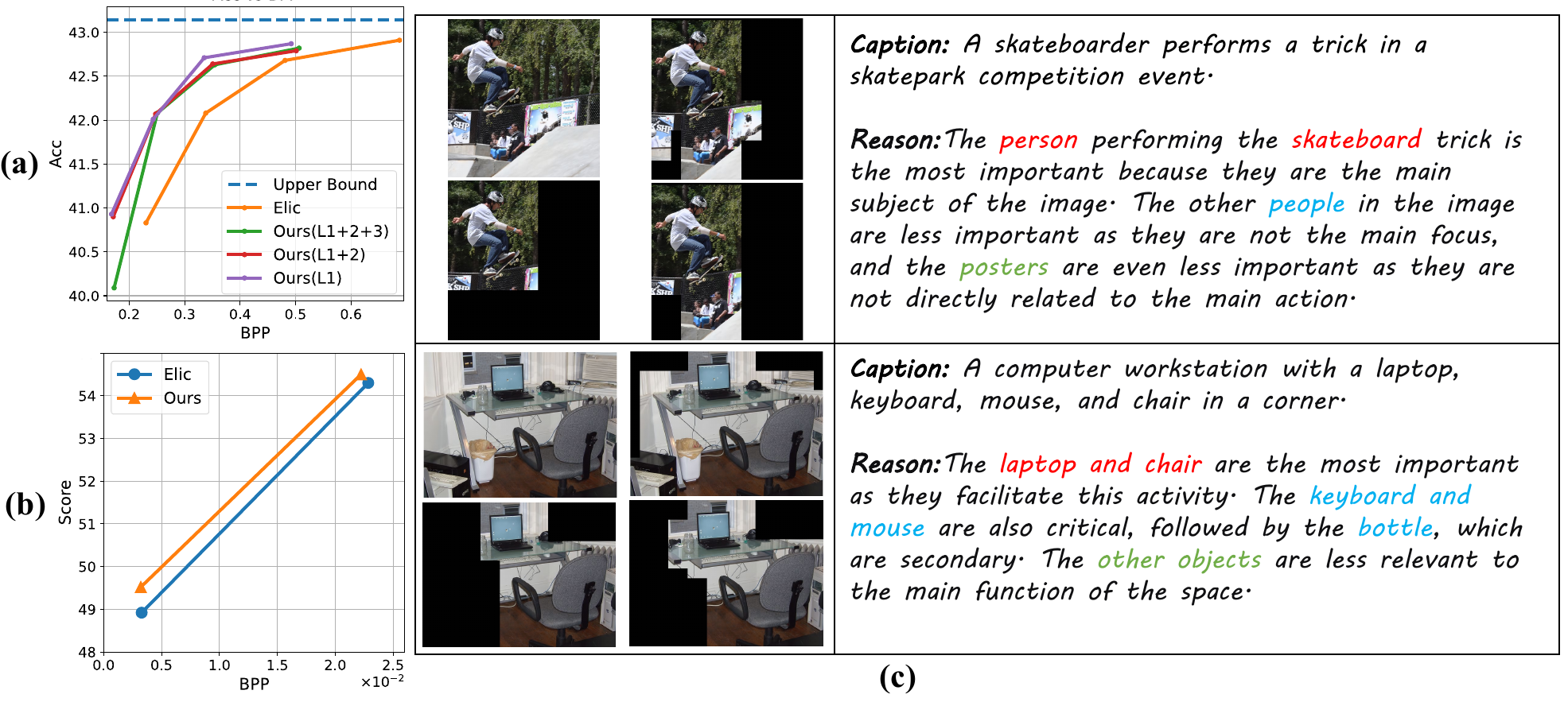}
    \vspace{-8mm}
    \caption{(a) Ablation studies on different importance. (b) VQA experiments. (c) The visualization (left) and interpretability (right)
 of SDComp. The transmitted bit-stream decreases in the clockwise direction, but still keep the same caption. LMM explains the rationale behind the ranks \textcolor{red}{$L_1$}, \textcolor{blue}{$L_2$}, \textcolor{green}{$L_3$}.}
 \vspace{-5mm}
    \label{fig:understanding}
\end{figure}

\vspace{-1mm}
We conduct ablation studies on image understanding using large models based on the Stanford-Cars dataset~\cite{KrauseStarkDengFei-Fei_3DRR2013}. As shown in Figure~\ref{fig:understanding}(a), we explore the impact of transmitting bitstreams of different importance levels on this task. It can be observed that transmitting all the objects extracted through grounding (i.e., $L_1$, $L_2$, and $L_3$) significantly improves performance compared with ELIC. Furthermore, leveraging the understanding capabilities of the LMM~\cite{chen2023internvl}, we can transmit only the more important bitstreams. Transmitting only the $L_1$ bitstream reduces the bitrate overhead, thereby enhancing the rate-distortion performance.

Moreover, we validate the effectiveness of SDComp in the image understanding task of visual question answering (VQA) on Vizwiz dataset~\cite{gurari2018vizwiz} by using LLaVA~\cite{liu2024visual} in Figure~\ref{fig:understanding}(b). SDComp achieves better rate-distortion performance.
\vspace{-2mm}
\subsection{Visualization Results}
\vspace{-1mm}
Figure~\ref{fig:understanding}(c) presents the visualization results with the varying importance levels identified by SDComp. We can see that SDComp retains significant objects while discarding less important ones. Even with the reduced number of transmitted objects, the overall understanding of the image remains largely unaffected. When utilizing an InternVL-1.5~\cite{chen2023internvl} for image captioning, the captions generated are the same as those from the original image. Furthermore, we prompt the InternVL-1.5 to output explanations for the ranks to explore the interpretability of the importance rank. The results demonstrate that the rank made by our method is indeed justifiable.
\vspace{-2mm}
\section{Conclusion}
\vspace{-1mm}
We propose a novel compression paradigm of Semantically Disentangled Compression (SDComp), which leverages the semantics comprehension ability of a large model to achieve ``intelligently coding for machine''. SDComp comprises three key components—visual grounding, content importance ranking, and structured image compression, they collaboratively compress and transmit objects accordingly based on their importance levels and task relevance. SDComp shows great superiority in rate-accuracy performance compared to traditional codecs, and allows partial decoding by retaining main content while discarding less important ones, maintaining overall semantic integrity. Experiments demonstrate the effectiveness and flexibility of SDComp in various downstream tasks like detection, segmentation, classification, VQA, and captioning.

\clearpage
\clearpage

\bibliographystyle{IEEEtran}
\bibliography{ref}
\end{document}